\documentclass[preprint,12pt]{clear2025} 

\usepackage{booktabs}
\usepackage{amsfonts,amsmath,amssymb}
\usepackage{bm}
\usepackage{siunitx}
\usepackage{algorithm}
\usepackage{algorithmic}

\usepackage{multirow}
\usepackage{makecell}

\usepackage{caption}
\usepackage{subcaption}

\title[Granger Causality, Gradient Descent and Pruning]{A Granger-Causal Perspective on Gradient Descent with Application to Pruning}
\usepackage{times}

\clearauthor{%
 \Name{Aditya Shah} \Email{aditysha@google.com}\\
 \addr Google Search, Google Austin, Texas, USA%
 \AND
 \Name{Aditya Challa} \Email{adityac@goa.bits-pilani.ac.in}\\
 \addr APPCAIR and CS\&IS, BITS Pilani KK Birla Goa Campus, India%
 \AND
 \Name{Sravan Danda} \Email{dandas@goa.bits-pilani.ac.in}\\
 \addr APPCAIR and CS\&IS, BITS Pilani KK Birla Goa Campus, India%
 \AND
 \Name{Archana Mathur} \Email{mathurarchana77@gmail.com}\\
 \addr Nitte Meenakshi Insitute of Technology, Yelahanka, Bangalore, India%
 \AND
 \Name{Snehanshu Saha} \Email{snehanshus@goa.bits-pilani.ac.in}\\
 \addr APPCAIR and CS\&IS, BITS Pilani KK Birla Goa Campus, India%
}

\def\eqref#1{equation~\ref{#1}}

\def\1{\bm{1}}

\def\vx{{\bm{x}}}

\DeclareMathAlphabet{\mathsfit}{\encodingdefault}{\sfdefault}{m}{sl}
\SetMathAlphabet{\mathsfit}{bold}{\encodingdefault}{\sfdefault}{bx}{n}

\newcommand{\pdata}{p_{{data}}}

\begin{document}

\maketitle

\begin{abstract}
    Stochastic Gradient Descent (SGD) is the main approach to optimizing neural networks. Several generalization properties of deep networks, such as convergence to a flatter minima, are believed to arise from SGD. This article explores the \emph{causality aspect of gradient descent}. Specifically, we show that the gradient descent procedure has an implicit granger-causal relationship between the reduction in loss and a change in parameters. By suitable modifications, we make this causal relationship explicit.

    A causal approach to gradient descent has many significant applications which allow greater control. In this article, we illustrate the significance of the causal approach using the application of \emph{Pruning}. 
    
    The causal approach to pruning has several interesting properties - (i) We observe a phase shift as the percentage of pruned parameters increase. Such phase shift is indicative of an optimal pruning strategy. (ii) After pruning, we see that minima becomes ``flatter'', explaining the increase in accuracy after pruning weights.
\end{abstract}


\section{Introduction}
\label{sec:intro}

Stochastic Gradient Descent (SGD) is the standard optimization procedure for deep neural networks. However, it surprised the researchers that SGD obtains solutions which generalize well. Several explanations such as sharp vs flat minima are given as evidence that the generalizability of deep networks is primarily due to properties of SGD \citep{DBLP:conf/nips/NeyshaburBMS17, DBLP:conf/icml/ZhuWYWM19}. In this article we explore the \emph{implicit causality in gradient descent procedure}.

\paragraph{Remark:} Throughout the article we use the word causality to refer to Granger-type causality and not causality in the sense of graphical models. 

The working hypothesis here is that -- Gradient descent implicitly (and inefficiently) performs causal reasoning. And, by making the causal relationship explicit one can obtain a refined model of learning. We evidence the above hypothesis by applying this idea to pruning. We observe that removing the \emph{non-causal} parameters almost always improves the accuracy and results in a much flatter minima.

\paragraph{Implicit causality in gradient descent:} Note that every step of gradient descent results in a change of both the parameters $\theta$ and the loss function $L$. The change in parameters is dictated by the gradient of the loss function $\partial L / \partial \theta$. The implicit causality within this model is that - Changing $\theta \to \theta + \Delta \theta$ is expected to change the loss $L \to L + \Delta L$, where $\Delta L$ and $\Delta \theta$ are related by the first-order gradient information. In other words, \emph{``changing the parameters $\theta$ \underline{causes} the change in the loss''}. This however is not always true - there usually are parameters which do not result in the reduction in the loss. Thus, the implicit causality in the gradient descent is not perfect. We make the causality relationship explicit in section~\ref{sec:2}. 

\subsection{Application to Pruning} 

Making the causal relationship explicit can have several advantages. In this article we illustrate the advantage of this model for pruning. 

\paragraph{Related Literature on Pruning:} Pruning neural networks has gathered a lot of attention for variety of reasons - (i) Improves the efficiency of models at inference without compromising the accuracy \citep{DBLP:conf/icml/KalchbrennerESN18, DBLP:journals/jmlr/HoeflerABDP21}, (ii) Lottery ticket hypothesis (LTH) \citep{DBLP:conf/iclr/FrankleC19,DBLP:conf/nips/JinC0FD22,DBLP:conf/iclr/PaulCLFGD23} and related works show that pruning experiments can help us understand the working of optimization and generalization in the training of neural networks.  Two broad approaches for pruning are: \textbf{(i) Magnitude Based Pruning:} The popular heuristic that, the magnitude of the parameter reflects the importance of the weight, underlies several existing pruning methods. \citep{DBLP:journals/corr/HanMD15,DBLP:conf/rep4nlp/GordonDA20,DBLP:journals/corr/abs-2301-05219}. In fact, as \citet{DBLP:journals/corr/abs-2301-05219} notes, using magnitude-based pruning at the level of filters (a.k.a $L_1$ norm pruning), the authors could achieve state-of-the-art results with this simple heuristic. \textbf{(ii) Impact Based Pruning:} Several pruning methods \citep{DBLP:conf/nips/CunDS89,DBLP:conf/nips/SinghA20} measure the dip in the loss function with respect to the parameters and decide which parameters to prune. Approaches such as one proposed by \citet{DBLP:conf/nips/SinghA20} use second-order Taylor approximation for the criterion. \citet{DBLP:conf/icml/BenbakiCM0P0M23} uses combinatorial optimization and approximates the Hessian using a low-rank matrix.

\begin{figure}[t]
\subfigure[][]{
\label{fig:1}
\includegraphics[width=0.45\linewidth]{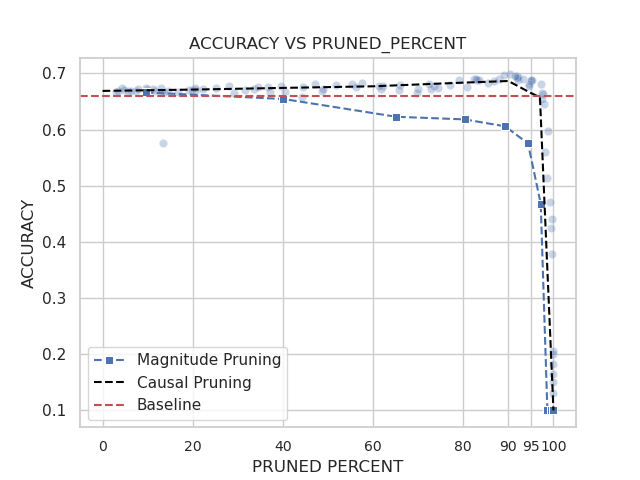}
}%
\hfill %
\subfigure[][]{
\label{fig:2}
\includegraphics[width=0.45\linewidth]{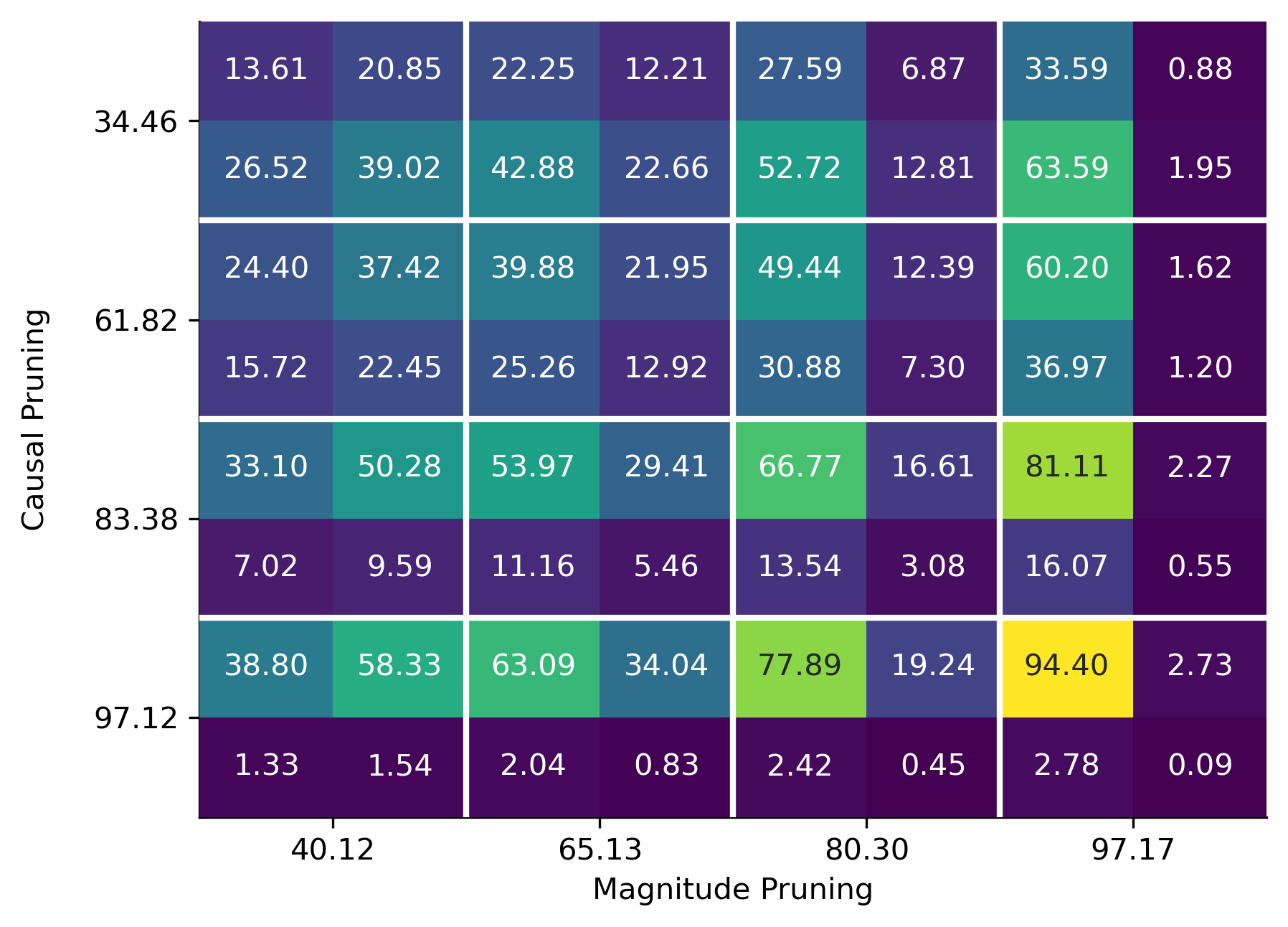}
}%
\caption{(a) Experiment on over-parameterized LeNet on CIFAR10 \emph{(Top-Right is better)}. There exists an accuracy-percent\_pruned trade-off in general. Causal pruning achieves a much sharper transition than magnitude pruning. This evidences that causal pruning identifies better weights to prune than magnitude pruning. (b) Comparing Masks of Causal Pruning vs Magnitude Pruning. We compare 4 different masks of causal pruning with 4 different masks of magnitude pruning. In each 2x2 matrix, - (0,0) entry shows the percentage of intersection between the masks, (1,0) entry shows \% of elements in magnitude pruning but not causal pruning, (0,1) entry shows the \% of elements in causal pruning but not magnitude pruning, and (1,1) entry shows the number of elements not pruned by either. Observe that there is a decent overlap between the causal pruning masks and magnitude pruning masks. The values on the axes show the total $\%$ pruned. The results presented in (a), together with those in (b), indicate that while magnitude pruning is a strong heuristic, it fails to consistently identify the optimal set.}
\end{figure}

\paragraph{Causal vs Magnitude Pruning -- Key results:} 

In this article, we prune the parameters which turn out to be non-causal and hence refer to this procedure as \emph{causal pruning}. As evidence to our hypothesis, we indeed observe that causal pruning finds a better optimal subset than the competitive methods such as magnitude pruning. Below we summarize the key results of which provides evidence to this claim. The details of the algorithm are delegated to the later sections.

The common approach to validate the pruning procedure is to observe the drop in test accuracy with respect to pruning\_percentage and measure the computation requirement using metrics such as FLOP count, etc. While these measures are suited for gauging the suitability of pruned network, they do not measure the optimality of the pruning procedure. We use two tests in this article, apart from test accuracy, to analyze optimality of the pruning procedure.

\paragraph{Test 1: A Phase Shift when plotting accuracy vs. pruning\_percent} It is known that there exists a trade-off between accuracy and percentage pruned. If a pruning procedure is optimal in perfectly identifying the subset of "unimportant" parameters, we expect a sharp decrease in accuracy, even with a small change in percentage pruned. However, we see a smoother transition if the approximation of the subset is sub-optimal. Figure~\ref{fig:1} shows a comparison of iterative versions of magnitude pruning and causal pruning. Intuitively, one does not expect much pruning in small networks (less than 1M parameters). Here we consider a slightly over-parameterized LeNet (with $657,080$ parameters) on the CIFAR10 dataset. Observe that the causal pruning reaches very close to the top-right corner, where removing even $\approx 2\%$ (going from 96\% to 98\% pruned\_percent) more parameters results in a significant drop in accuracy. Contrast this with magnitude pruning, where the drop is much ``smoother". Thus this provides evidence that causal pruning can identify the parameters to be pruned much better than the magnitude heuristic.

Figure~\ref{fig:2} compares the masks obtained by magnitude and causal pruning. Observe a decent overlap between the masks obtained by magnitude and causal pruning, indicating a strong correlation. However, at lower values of $\%$ pruned, the two methods diverge. The interesting case to consider is when both magnitude and causal pruning prune $\approx 97.1\%$ of the weights, the overlap is $94.4\%$. However, causal pruning attains an accuracy of $68.11\%$ while magnitude pruning attains $46.8\%$. Surprisingly, these masks only differ by $2.7\%$ of the weights. This shows that even minor differences in the weights pruned can lead to substantially different results.

\paragraph{Test 2: Flatter Minima} It is widely believed that sharp minima lead to bad generalization and flat minima lead to better generalization \citep{DBLP:conf/iclr/JiangNMKB20}. If the pruning procedure is optimal, i.e., it reduces the number of parameters without affecting the accuracy, the model complexity is expected to reduce and we expect a flatter minima. Moreover, this also explains why few pruning methods seem to \emph{increase} the accuracy. We see that causal pruning obtains a much flatter minima than magnitude pruning. This is discussed in detail in section~\ref{sec:exp}.

\section{Causality and Gradient Descent}
\label{sec:2}

\paragraph{Notation:}  Let $\pdata=\{\vx_i, y_i\}$ denote the dataset. Let $f_{\theta}$ denote the network to be trained. Let $L(\theta)$ denote the loss function used to optimize $\theta$ using Stochastic Gradient Descent (SGD). We use $\partial L (\theta^t)/\partial \theta$ to denote the derivative of $L$ with respect to $\theta$ at $\theta=\theta^t$. 

Let $\theta^0, \theta^1, \cdots, \theta^{t}, \cdots$ denote the path in the parameter space taken by the gradient descent. Further, we let
\begin{equation}
    (\Delta L)^t = L(\theta^{t}) - L(\theta^{t-1})
\end{equation}
Let $\theta^{t}_i$ denote the $i^{th}$ parameter in the vector $\theta^{t}$. Then, we denote,
\begin{equation}
    (\Delta \theta_{k})^t = \theta^{t}_{k} - \theta^{t-1}_{k}
\end{equation}

\paragraph{Vanilla Gradient Descent:}We have,
\begin{equation}
    \theta^{t+1} = \theta^{t} - \eta\frac{\partial L (\theta^{t})}{\partial \theta}
    \label{eq:1}
\end{equation}
where $\eta$ denotes a fixed learning rate. Now, using a first-order Taylor approximation of $L$, we have
\begin{equation}
    \begin{aligned}
        L(\theta^{t+1}) &= L(\theta^{t} - \eta\frac{\partial L (\theta^{t})}{\partial \theta}) \\
        & = L(\theta^{t}) - \eta \left(\frac{\partial L (\theta^{t})}{\partial \theta}\right)^T \frac{\partial L (\theta^{t})}{\partial \theta}
    \end{aligned}
\end{equation}
Substituting from \eqref{eq:1},
\begin{equation}
\label{eq:2}
    \begin{aligned}
        L(\theta^{t+1}) & = L(\theta^{t}) - \eta \left(\frac{\theta^t - \theta^{t+1}}{\eta}\right)^T \left(\frac{\theta^t - \theta^{t+1}}{\eta}\right)\\
        & =  L(\theta^{t}) - \frac{1}{\eta} \|\theta^t - \theta^{t+1}\|^2 \\
        & =   L(\theta^{t}) - \frac{1}{\eta} \sum_k (\theta^{t}_k - \theta^{t+1}_k)^2 \\
    \end{aligned}    
\end{equation}

\paragraph{Granger Causal Interpretation:} \eqref{eq:2} can be interpreted as a specific case of vector-based Granger causality where the time-lag is considered to be $1$ when written as below:
\begin{equation}
    L^{t+1} = L^{t} + \sum_{k} \gamma_{k} (\theta^{t+1}_{k} - \theta^{t}_{k})^2
\end{equation}
where $\gamma_k = - 1/\eta$ for all $k$. In words, \eqref{eq:2} can be interpreted as -- \emph{``Changing the parameter value by $\Delta \theta$ would \underline{cause} the loss to reduce by the value $\Delta L$.''} Nevertheless, this is not true in general. The loss does not reduce by $\Delta L$ but rather reduces by an unknown quantity. So, while the gradient descent assumes \emph{implicit causal relationship} between the parameters and the loss reduction, it does not verify it nor does it adjust accordingly. In what follows we shall make this implicit causal relation explicit.

\paragraph{Remark:} Note that we have only considered vanilla gradient descent above. Extending this analysis to SGD+momentum approaches and possibly SGD+momentum+adaptive-learning rates will result in a more comprehensive granger-casual model. A brief discussion of this can be found in appendix~\ref{sec:sgd_momentum}. In this article, we are only interested in the model for vanilla gradient descent. 

\paragraph{Making causal relation explicit by replacing $1/\eta$ with $\gamma_k$:} 
Now generalizing from \eqref{eq:2}, assuming that each $\theta^{t}_k$ does not contribute equally to the reduction in the loss, we propose the following linear model which captures the dynamics of gradient descent -
\begin{equation}
    \Delta L = \sum_{k} \gamma_k (\Delta \theta_k)^{2}
    \label{eq:vanillasgdmodel}
\end{equation}
Here $\gamma_k$ denotes the importance of parameter $\theta_k$. This is in line with the Granger-causal interpretation above.

By replacing the constant learning rate across all parameters with parameter specific $\gamma_k$, we are able to capture information that is not possible to measure otherwise. As we shall exhibit in this article, this information is crucial for identifying the importance of a parameter. Specifically, we have the following principle -- \emph{Consider the parameter $\theta_k$ to be important if $\gamma_k \neq 0$, and unimportant if $\gamma_k = 0$.}

\begin{figure}[t]
\centering
\subfigure[][]{%
    \label{fig:3a}
    \includegraphics[width=0.3\linewidth]{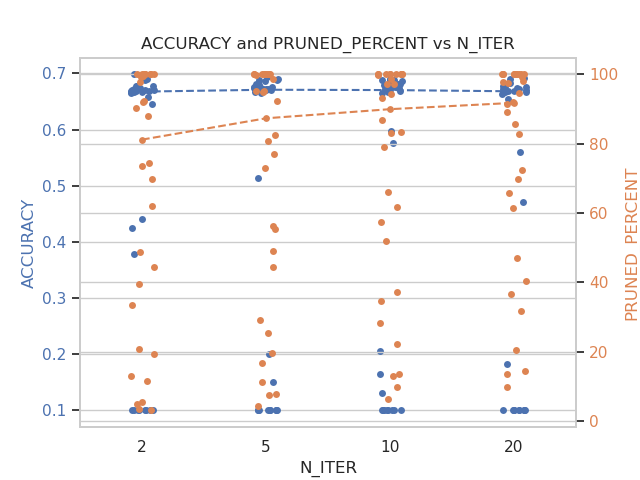} %
} %
\hfill%
\subfigure[][]{%
    \label{fig:3b}
    \includegraphics[width=0.3\linewidth]{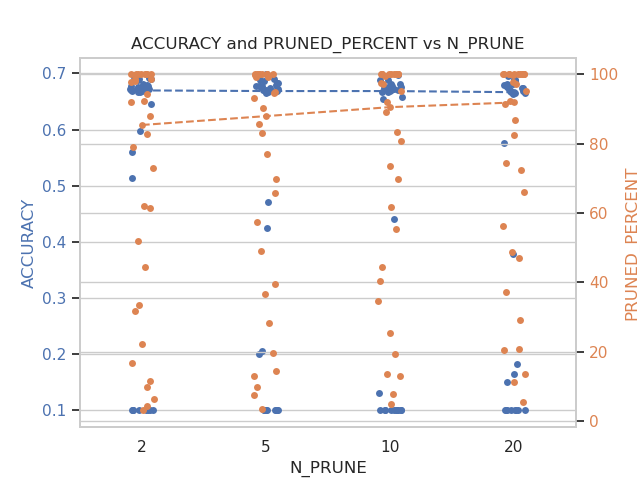}
    } %
\hfill %
\subfigure[][]{
    \label{fig:3c}
    \includegraphics[width=0.3\linewidth]{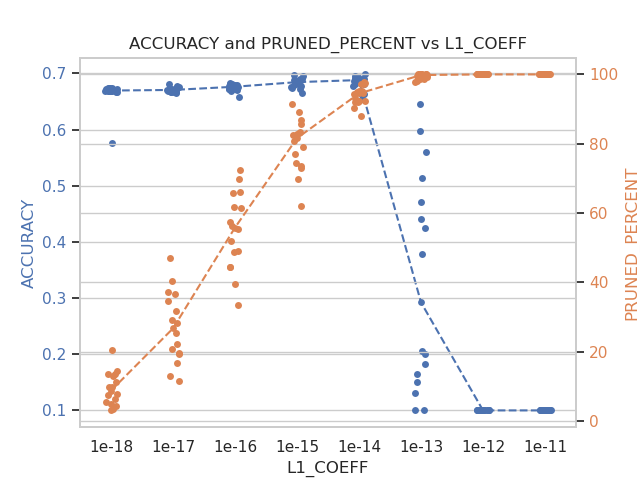}
}
\caption{Effect of hyperparameters $N_{iter}$, $N_{prune}$ and \texttt{L1\_coeff}. These experiments have been performed using LeNet on CIFAR10. The dotted lines indicate the median percentage pruned and accuracy for each value.   Observe that accuracy remains robust to the changes in $N_{iter}$ and $N_{prune}$, while percentage pruned increases as $N_{iter}$ and $N_{prune}$ increase. We suggest default values of $N_{iter}=10$ and $N_{prune}=10$ as a balance between the computational complexity and pruning efficiency. \texttt{L1\_coeff} is the most important parameter which decides the percentage pruned. Interestingly, we observe that adding more parameters decreases the accuracy. This is explained by the observation that causal pruning reaches a much flatter minima than the unpruned version.}
\label{fig:3}
\end{figure}

\section{Causal Pruning Algorithm}

\begin{table*}[ht]
\caption{List of parameters}
\centering
\begin{tabular}{p{0.15\linewidth}p{0.5\linewidth}p{0.2\linewidth}} 
\toprule
Name & Explanation & Typical Values \\
\midrule 
$N_{pre}$ & Number of epochs for training the network after which the pruning is performed & 5-10\\
$N_{iter}$ & Number of iterations of pruning to be performed & 2-10\\
$N_{prune}$ & Number of epochs for training the network to collect the data required for causal pruning & 2-10\\
$N_{post}$ & Number of epochs for training the network after pruning to evaluate the performance & 100-300\\
\texttt{L1\_coeff} & The regularization parameter used for \texttt{LassoRegression} & \num{1e-18} to \num{1e-11} (log space)\\
\bottomrule
\end{tabular}
\label{table:1}
\end{table*}

\begin{algorithm}[t]
\caption{Causal Pruning}\label{alg:causal pruning}
\begin{algorithmic}[1]
\STATE \textbf{Parameters:} $N_{pre}$, $N_{iter}$, $N_{prune}$, $N_{post}$, \texttt{L1\_coeff}. (See table~\ref{table:1} for details).
\STATE \textbf{Input:} Model:$f_{\theta}$, Dataset:$\{\vx_i, y_i\}$
\STATE Train the model for $N_{pre}$ number of epochs. \label{alg:stepa}
\FOR{$i=1,2,\cdots,N_{iter}$}
    \STATE Train the model for $N_{prune}$ epochs and collect the values of the parameters and the losses after each gradient step. Let $\{(\theta^t, L^t)\}$ denote the values after each gradient step. 
    \STATE For each layer, using \texttt{Lasso Regression}, fit the model in  \eqref{eq:vanillasgdmodel} using the $L_1$ coefficient \texttt{L1\_coeff}. Prune the parameters for which $\gamma_k=0$.
    \STATE Reset the remaining weights to be the ones after step \ref{alg:stepa}.
\ENDFOR
\STATE Complete the training of the model for $N_{post}$ number of parameters.
\end{algorithmic}
\end{algorithm}

\paragraph{Pruning as Feature Selection:} From \eqref{eq:vanillasgdmodel}, we have a linear model with features/independent variables $\{(\Delta \theta_k)^2\}$ with coefficients $\gamma_k$ and dependent variable $\Delta L$. Thus, selecting the parameters is the same as testing $H_0:\gamma_k=0$ vs $H_1:\gamma_k \neq 0$. In other words, selecting a subset of parameters (a.k.a pruning) is the same as finding the subset of features in the above linear model. This matches the Granger-causal interpretation of causal parameters as well. A vast amount of literature from classical statistics \citep{James2023} deals with this problem. We use a sparse solution based on $L_1$ regularization for subset selection. 

\paragraph{Pretraining:} Given a dataset and architecture, we train the model for $N_{pre}$ epochs. This is crucial as pointed out by \citet{DBLP:conf/mlsys/BlalockOFG20}. This phase decides the \emph{basin of attraction} of the parameters after which the model can be pruned more effectively. 

\paragraph{Causal Pruning:} After the pretraining, we have the pruning phase. Here, we further train the model for $N_{prune}$ epochs and collect the parameters and losses after each gradient step. Let $\{\theta^{t}, L^{t}\}_{t=0}^{T}$ denote the values we obtain. Starting at $t=1$, compute first-order differences,
\begin{equation}
    \begin{aligned}
        \Delta \theta^{t} &=\theta^{t} - \theta^{t-1} \\
        \Delta L^{t} &= L^{t} - L^{t-1}
    \end{aligned}
\end{equation}
and fit the model using $\{\Delta \theta^{t},\Delta L^{t}\}_{t=1}^{T}$ as the dataset. Here $\Delta \theta$ denotes the independent variables and $\Delta L$ denotes the dependent variables.
\begin{equation}
    \Delta L = \sum_{k} \gamma_k (\Delta \theta_k)^{2} + \alpha \sum_k |\gamma_k|
    \label{eq:lasso_vanillasgd}
\end{equation}
where $\alpha$ denotes the $L_1$ regularization coefficient. We obtain the pruning masks as $\{\theta_k \mid \gamma_k \neq 0\}$. Repeat this process $N_{iter}$ times to get the final mask. After each pruning step, we \emph{rewind} the weights to the weights obtained after pretraining.  This allows for pruning a larger subset of weights. 

\paragraph{Post Training:} To maximize the performance of the network, we train it further (using only unpruned weights) for $N_{post}$ number of epochs. 

\subsection*{Important Remarks:}

\paragraph{Optimization of \eqref{eq:lasso_vanillasgd}:} For some networks, the number of parameters within each layer can go up to $16M$ and hence can cause memory issues. \citet{DBLP:journals/jmlr/Shalev-ShwartzT11} discusses several algorithms for stochastic optimization of \eqref{eq:lasso_vanillasgd}. We follow the procedure outlined in \citep{DBLP:conf/acl/TsuruokaTA09}. We also further restrict the feature (parameter) selection with each layer, i.e. solve \eqref{eq:lasso_vanillasgd} for each layer separately.  

\paragraph{Computational Complexity:} Apart from the usual training of the networks, this procedure requires two additional steps - (i) Saving the checkpoints of the model after each gradient step for a few epochs, and (ii) Solving the lasso regression of the \eqref{eq:lasso_vanillasgd}. Note that saving the checkpoints does not increase the computational complexity, but does require a larger storage. Further, these checkpoints can be removed after the computation of the coefficients in \eqref{eq:lasso_vanillasgd}. Solving \eqref{eq:lasso_vanillasgd} is a straightforward problem and there exist several efficient solutions which use stochastic gradients \citep{DBLP:journals/jmlr/Shalev-ShwartzT11,DBLP:conf/acl/TsuruokaTA09}. The naive approach has the computational complexity of $\mathcal{O}(mdk)$ where $m$ is the number of samples, $d$ is the number of features and $k$ is the number of epochs required to reach the solution. Note that it scales linearly in both the number of samples and some parameters (features), and hence very efficient. Further, these algorithms can also be parallelized, leading to higher gains. 

\paragraph{Parameter-Level or Layer-Level or Entire-Network:} It sometimes helps to prune channels or layers instead of individual parameters. This is referred to as structured pruning \citep{DBLP:journals/pami/HeX24}. One can adapt the procedure in \eqref{eq:vanillasgdmodel} to structured pruning using the strategy of \emph{weight-sharing}. In essence, enforce the constraint that $\gamma_k$ is equal for all parameters $\theta_k$ within a layer. In this article, we only consider Parameter-Level pruning a.k.a unstructured pruning.

\paragraph{Difference from existing Magnitude/Impact based pruning:} The proposed pruning method is substantially different from existing approaches (see section~\ref{sec:intro}). Magnitude-based pruning methods rely heavily on the magnitude heuristic, while we consider the gradient descent dynamics. Impact-based pruning methods estimate the dip in the loss with respect to each parameter, which is used for pruning. However, we on the other hand compare the predicted dip in the loss with the actual dip in the loss and use this for defining and pruning unimportant parameters.

\begin{figure}[t]
\centering
\subfigure[LeNet-CIFAR10][]{%
    \label{fig:4a}
    \includegraphics[trim={0 0 0 1.7cm},clip,width=0.45\linewidth]{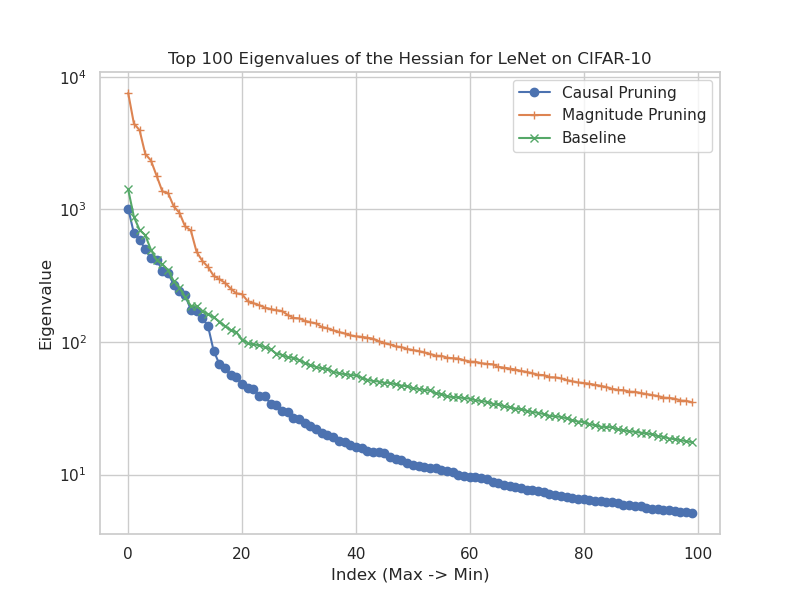}
}%
\hfill %
\subfigure[ResNet-CIFAR10][]{
    \label{fig:4b}
    \includegraphics[trim={0 0 0 1.3cm},clip,width=0.45\linewidth]{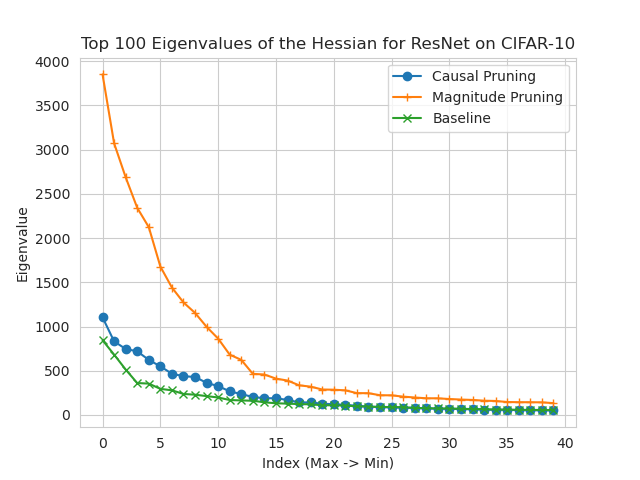}    
}
\caption{Top Eigenvalues at minima of various pruning methods. Understanding the flatness of the minima is the key to understanding the generalization of neural networks. With the ``right'' pruning approach we expect to see a flatter minima comparable to the original network. As shown in the plots, magnitude pruning tends to increase the eigenvalues (leading to sharper minima) compared to causal pruning (which gives flatter minima).   }
\label{fig:4}
\end{figure}

\begin{figure}[t]
\subfigure[Causal Pruning][]{%
    \label{fig:5d}
    \includegraphics[width=0.3\linewidth]{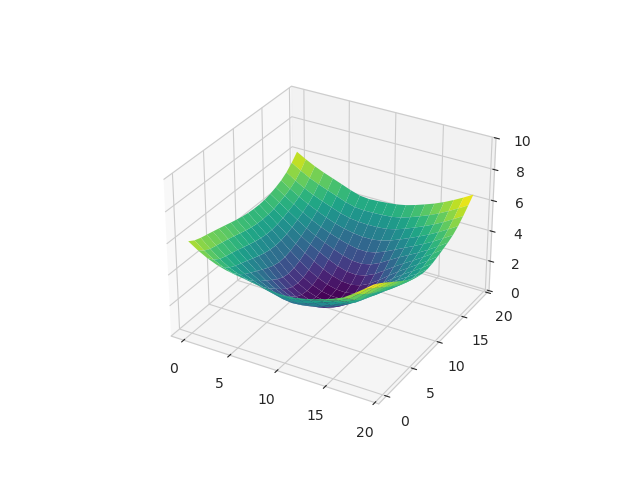}
}%
\hfill%
\subfigure[Magnitude Pruning][]{%
    \label{fig:5e}
    \includegraphics[width=0.3\linewidth]{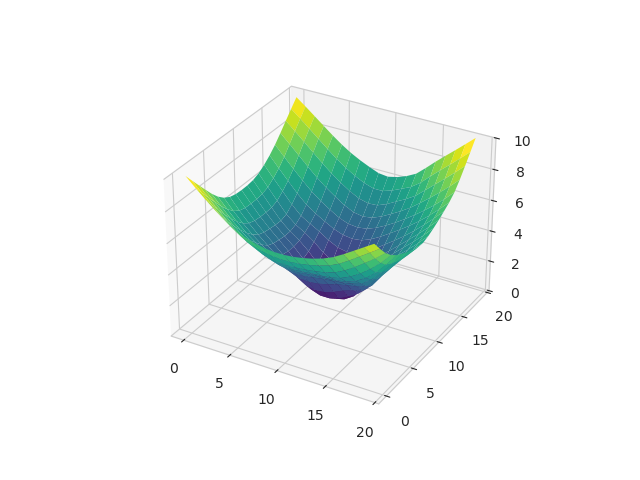} 
}%
\hfill%
\subfigure[No Pruning][]{%
    \label{fig:5f}
    \includegraphics[width=0.3\linewidth]{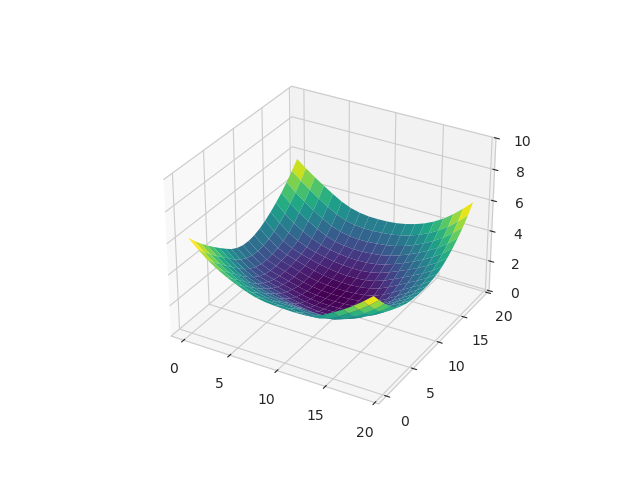}
}
\caption{Loss Landscape of ResNet18 on CIFAR10 for various pruning strategies. As one can see, causal pruning obtains a flatter minima close to the original network, while magnitude pruning increases the sharpness of the minima. This is another piece of evidence to support that - While magnitude pruning is a strong heuristic, it does prune a few ``important'' parameters. Figure~\ref{fig:a1} in the appendix shows the corresponding plots for LeNet on CIFAR10. }
\label{fig:5}
\end{figure}

\begin{table*}[h]
\caption{Comparing Causal Pruning with Magnitude pruning. The results are presented as percent-pruned (test accuracy).}
\centering
\begin{tabular}{ccccc}
{\makecell{Model \\ (\# Params)}}  & Dataset & No-Pruning & Magnitude Pruning & Causal Pruning \\
\toprule
\multirow{3}{*}{\makecell{LeNet \\ (657,080)}} & FashionMNIST & 0.0\%(90.51\%) & 65.13\%(90.31\%) & 68.72\%(90.39\%) \\
& CIFAR10 & 0.0\%(64.49\%) & 95.99\%(55.71\%) & 96.30\%(67.22\%)  \\
& TinyImageNet & 0.0\%(17.79\%) & 95.99\%(9.93\%) & 96.08\%(19.49\%) \\
\midrule
 \multirow{3}{*}{\makecell{AlexNet \\ (57,044,810)}}&  FashionMNIST & 0.0\%(91.05\%) & 95.99\%(91.09\%) & 93.37\%(91.25\%) \\
 &  CIFAR10 & 0.0\%(70.29\%) & 95.99\%(69.88\%) & 99.13\%(72.01\%) \\
 &  TinyImageNet & 0.0\%(37.38\%) & 88.71\%(41.89\%) & 88.72\%(41.64\%) \\
\midrule
\multirow{3}{*}{\makecell{ResNet18 \\ (11M)}} & FashionMNIST & 0.0\%(92.79\%) & 85.22\%(92.08\%) & 85.21\%(92.32\%) \\
& CIFAR10 & 0.0\%(74.05\%) & 91.11\%(76.18\%) & 91.17\%(78.45\%) \\
& TinyImageNet & 0.0\%(29.40\%) & 77.87\%(28.22\%) & 73.93\%(29.72\%) \\
\bottomrule
\end{tabular}
\label{table:2}
\end{table*}


\section{Experiments and Analysis:}
\label{sec:exp}

\paragraph{Scope of Experiments:} Recall that the key idea in this article is to make explicit the implicit causal relation within gradient descent. We consider the application of pruning to explore the power of this observation. Our focus is not on achieving state-of-the-art results but on substantiating the claim above. Specifically, causal pruning prunes an \emph{optimal} subset compared to the magnitude heuristic as a baseline. The code for these experiments can be found at \url{https://anonymous.4open.science/r/causalpruning-9DB5/README.md}



\subsection{Dependence on the Hyperparameters} The parameters required for causal pruning are shown in table \ref{table:1}. We consider $N_{pre}$ large enough that the SGD is within a basin of attraction, and $N_{post}$ large enough that there is no further improvement to the loss function. Hence, we only consider $N_{iter}$, $N_{prune}$ and \texttt{L1\_coeff}. Figure \ref{fig:3} shows the accuracy and percentage pruned for various values of the parameters. These studies have been done using LeNet on the CIFAR10 dataset. 
\paragraph{Accuracy of Causal pruning is fairly robust to $N_{iter}$ and $N_{prune}$:} Figure~\ref{fig:3a} shows the stripplot for various values of $N_{iter}$ and figure~\ref{fig:3b} shows the stripplot for $N_{prune}$. Recall that $N_{iter}$ denotes the number of times we repeat the pruning procedure, and $N_{prune}$ indicates the number of epochs we train the network to collect the data used for fitting \eqref{eq:lasso_vanillasgd}. The dotted lines indicate the median percentage pruned and accuracy for each value. 

Observe that the accuracy is unaffected at different values of $N_{iter}$ and $N_{prune}$. Again, this can be explained as the manifestation of our hypothesis that causal pruning identifies the important weights precisely. If this had not been the case, we expect changes in accuracy when changing the number of pruning steps. However, the percentage pruned increases nominally and we increase the values of $N_{iter}$ and $N_{prune}$. Increasing the values $N_{prune}$ and $N_{iter}$ from $10$ to $20$ improves the results nominally but at $2\times$ the cost. Hence, we consider $10$ the default value for these hyperparameters. 

\paragraph{Effect of \texttt{L1\_coeff}:} \texttt{L1\_coeff} is arguably the most important hyper-parameter which directly controls the sparsity. Figure~\ref{fig:3c} shows the changes in accuracy and percent\_pruned by changes in \texttt{L1\_coeff}. Concerning pruning\_percent, we observe the expected trend - as \texttt{L1\_coeff} increases, the percentage pruned also increases. This is expected since the \texttt{L1\_coeff} directly affects the number of $0$s in the solution of \eqref{eq:lasso_vanillasgd}, which in turn is used to prune the parameters. 

However, the interesting thing to note is the change in accuracy. Observe that at $\texttt{L1\_coeff}=$\num{1e-14} ($68.8\%$) has a (slightly) larger accuracy than $\texttt{L1\_coeff}=$\num{1e-18} ($66.9\%$). This is because causal pruning obtains a flatter minima than the unpruned model. Hence, adding the parameters leads to noise that results in slightly lesser accuracy. Also, as observed earlier, there is a sharp increase in the accuracy as $\texttt{L1\_coeff}$ changes from \num{1e-12} to \num{1e-14}.

An important deviation of causal pruning from the existing approaches is that we do not control the amount of pruning. Rather, we let the model decide the amount to be pruned based on the \texttt{L1\_coeff}.   

\subsection{Causal Pruning Obtains a Flatter Minima}

\paragraph{Review of Flat Minima vs Generalization:} One popular hypothesis to explain the generalization of deep neural networks is the ideas of \emph{flat minima}. Putting this simply, when the minima are flatter, one expects to see better generalization. \citet{DBLP:conf/iclr/JiangNMKB20} performs extensive experiments and shows that the \emph{sharpness of the minima} is the most correlated measure with generalization. \citet{DBLP:conf/iclr/ForetKMN21} proposes a sharpness-aware minimization optimizer, which is shown to have better results than the rest of the minima. \citet{DBLP:conf/icml/AhnJS24} uses the trace of the Hessian normalized by the dimension to measure the sharpness of the minima.

\paragraph{Flat Minima in the context of Pruning:} Another way to judge if a particular subset of parameters is the ``right'' subset to prune is that \emph{It should result in a flatter minima comparable to the original network.} The intuition behind this is, that as we reduce the number of parameters, the complexity of the model reduces, hence the minima should be flatter leading to a better/comparable generalization. Note that the flat minima perspective also explains why one sees a slight improvement in test accuracy with mild pruning. 

Figure \ref{fig:4} shows the top eigenvalues obtained with different pruning strategies for ResNet18 and LeNet on CIFAR10. We observe that causal pruning either improves (in the sense of flatter minima) the eigenspectrum of the Hessian or remains the same as the no-prune network. In contrast, magnitude pruning increases the eigenvalue spectrum by a large margin. Figure \ref{fig:5} visualizes the loss landscape of the minima obtained for ResNet18 on CIFAR10. Figure~\ref{fig:a1} (supplementary material) shows the corresponding plots for LeNet on CIFAR10. Details of these experiments and approaches used to generate these plots are discussed in appendix \ref{sec:details_flat_mimima}.

\paragraph{Comparison on other dataset/model combinations:} Table~\ref{table:2} shows the results across various dataset/model combinations. We consider 3 architectures - LeNet \citep{DBLP:journals/pieee/LeCunBBH98}, Alexnet \citep{DBLP:conf/nips/KrizhevskySH12} and ResNet18 \citep{DBLP:conf/cvpr/HeZRS16}, and 3 different datasets - FashionMNIST \citep{DBLP:journals/corr/abs-1708-07747}, CIFAR10 \citep{cifar10} and TinyImageNet \citep{tinyimagenet}.  Note that we outperform magnitude pruning in almost all these settings. As stated before, this is attributed to the ability of causal pruning to capture the dynamics of gradient descent. 

\emph{Comparison with CHITA \citep{DBLP:conf/icml/BenbakiCM0P0M23}:} While our primary goal is not to achieve state-of-the-art results, we present a comparison with the current leading method, CHITA, to provide a comprehensive perspective. Using ResNet50 on the ImageNet dataset for all experiments, we observe notable differences in performance within the context of single-shot pruning. Our causal pruning approach achieves an accuracy of $71\%$\footnote{$54\%$ after pruning and $71\%$ after fine-tuning}  with $80\%$ pruning, compared to CHITA's approximate $45\%$ with $80\%$ pruning\footnote{These figures are derived from Figure 2(a) in \citep{DBLP:conf/icml/BenbakiCM0P0M23} based on observation}. However, when employing gradual pruning (specifically, two pruning steps in our causal pruning process), our method reaches $56\%$ accuracy at $95\%$ pruning, whereas \citet{DBLP:conf/icml/BenbakiCM0P0M23} reports a $73\%$ accuracy at $95\%$ pruning.

It's important to note that our objectives differ significantly; we focus on understanding the intricacies and artifacts of pruning, whereas \citet{DBLP:conf/icml/BenbakiCM0P0M23} is geared towards optimizing pruning for speed and state-of-the-art performance. Additionally, these comparisons are observational, conducted without hyper-parameter tuning, and the computational setups\footnote{CHITA uses a batch size of 5000 and uses approx. 200 CPUs and 10 GPUs, while we perform the experiments on 40 CPUs and 4 GPUs.} involved vary significantly.

\section{Conclusion and Future Work}

In conclusion, the key idea in this article is to re-interpret gradient descent from the perspective of causality. Specifically, we make the Granger-causal perspective explicit by suitably modifying the gradient descent. This allows for a different framework for optimizing deep networks. We explored the application of pruning in this article. 

\paragraph{Causality and Pruning:} On pruning, the following are the key takeaways from this article:
\begin{itemize}
    \item When making the causal relationship explicit, we observe that each parameter is mapped to a $\gamma_k$, which holds a piece of important information. Specifically, whether a parameter is important or not can be decided by testing whether $\gamma_k = 0$ vs $\gamma_k \neq 0$. That is, by testing whether the corresponding $\theta_k$ is causal or not. Thus, we refer to this technique as \emph{causal pruning}.
    \item To implement causal pruning, we frame the problem of testing causality as a lasso regression problem. Thanks to the vast literature on lasso regression, this can be implemented very efficiently.
    \item \emph{Judging if a specific subset of parameters are the right parameters to be pruned?} Traditionally test accuracy has been the approach to validate whether a pruning procedure is optimal. In this article, we propose and use 2 more validation techniques 
    \begin{itemize}
         \item \textbf{Phase Shift Validation:} The pruning procedure should result in a \emph{phase-shift} when plotting accuracy vs percent\_pruned. The sharper the transition, the better the method. Intuitively this is because, if there exists a perfect subset of parameters for pruning, then pruning any more parameters above this should reduce the accuracy drastically.
         \item \textbf{Flat Minima Validation:} An optimal pruning procedure should not increase the sharpness of minima. Specifically, it should not increase the large eigenvalues leading to a different output from the original unpruned network. Moreover, this also explains why few pruning approaches tend to increase test accuracy.
    \end{itemize}   
    \item We show that causal pruning is better than the currently widely used approach of magnitude pruning on all these measures.
\end{itemize}

\paragraph{Extensions to Causal Pruning:} There are several possible extensions to the present work (i) In this article we focus on unstructured pruning. However, a simple extension to structured pruning is possible by using a weight-sharing strategy in \eqref{eq:lasso_vanillasgd}, where the lasso coefficients are shared as per the structure. For instance, let $\theta_k$ denote each parameter and $\gamma_k$ represent the corresponding lasso coefficient. To prune entire layers/filters, ensure that $\gamma_i = \gamma_j$ for any parameters $\theta_i$ and $\theta_j$ that belong to the same layer/filter. This uniformity ensures that parameters within a single layer are treated equivalently during the pruning process. (ii) One can extend the lasso model in \eqref{eq:lasso_vanillasgd} to SGD with momentum and even to SGD with momentum and adaptive learning rates. This is briefly discussed in the supplementary material (appendix \ref{sec:sgd_momentum}). (iii) In the current manuscript, we have performed a layer-wise optimization of \eqref{eq:lasso_vanillasgd}. We intend to extend this to a single optimization for the entire network, from which we expect modest gains. 

\paragraph{Other Applications of Causal Interpretation:} We envision several other applications to the causal perspective of gradient descent.
\begin{itemize}
    \item \emph{Causal Second Order Optimization:} A slightly different interpretation of the causal pruning presented in this article is -- We perform \emph{two first-order optimizations which can encapsulate second-order information.} Specifically, we perform one first-order optimization on reducing the loss, and the second first-order optimization based on the explainability (causality) of parameters to the reduction of the loss. This, in effect, is equivalent to the second-order optimization method. Specifically, we believe that the parameters $\gamma_k$ obtained as causal pruning are related to the ``optimal'' parameter-specific learning rate of the first-order optimization. We aim to explore this as part of the future work.
    \item \emph{Increasing the capacity of the network using causality:} Note that pruning is aimed at reducing the size of the model for efficient inference. However, a significant drawback of these approaches is that -- Training still requires one to consider the entire model. And since training the model is arguably the costliest aspect, the impact of pruning on cost is rather small.

    Instead, one can ask the question -- \emph{Can we start with a small model and increase its capacity at the time of optimization?} This article provides one approach to answer this question in the affirmative. Specifically, at the time of optimization, if a specific parameter turns out to be redundant, one can reinitialize the parameter, which increases the capacity. For such an approach to work, the criterion selected for pruning should be optimal -- i.e. reduce both false positives and false negatives. As we show in this article, \emph{causal pruning} satisfies this objective. Hence, as part of future work, we aim to explore this line of research as well.
\end{itemize}

\acks{Aditya Challa acknowledges the support from CEFIPRA(68T05-1). Snehanshu Saha and Aditya Challa would like to thank the Anuradha and Prashanth Palakurthi Center for Artificial Intelligence Research (APPCAIR) and SERB CRG-DST (CRG/2023/003210) for their support. Snehanshu Saha acknowledges SERB SURE-DST (SUR/2022/001965) and the DBT-Builder project (BT/INF/22/SP42543/2021), Govt. of India for partial support. Archana Mathur would like to thank SERB TARE (TAR/2021/000206) for their support}

\bibliography{mybibfile}

\appendix

\section{Deriving the Causal Relation in the case of SGD + Momentum}
\label{sec:sgd_momentum}

Let's consider the following variant of the gradient descent with momentum-
\begin{equation}
\label{eq:3a}
\begin{aligned}
    v^{t+1} &= \beta v^{t} + \frac{\partial L(\theta^t)}{\partial \theta} \\
    \theta^{t+1} &= \theta^t - \eta v^{t+1}
\end{aligned}    
\end{equation}
where $\beta$ denotes the momentum hyperparameter. Then, we have, using first-order Taylor approximation,
\begin{equation}
\label{eq:3b}
\begin{aligned}
    L(\theta^{t+1}) &= L(\theta^t) -  \eta \left(\frac{\partial L(\theta^t)}{\partial \theta}\right)^Tv^{t+1}\\
    &= L(\theta^t) - \eta \beta \left(\frac{\partial L(\theta^t)}{\partial \theta}\right)^Tv^{t} - \eta \left(\frac{\partial L(\theta^t)}{\partial \theta}\right)^T\frac{\partial L(\theta^t)}{\partial \theta}
\end{aligned}    
\end{equation}
Now from \eqref{eq:3a} using previous time steps, we can derive
\begin{equation}
    v^{t} = \frac{\theta^{t} - \theta^{t-1}}{\eta} = \frac{\Delta \theta^t}{\eta}
\label{eq:4a}
\end{equation}
and,
\begin{equation}
    \frac{\partial L(\theta^t)}{\partial \theta} = -\frac{v^{t+1} - \beta v^t}{\eta}  = -\frac{\Delta \theta^{t+1} - \beta \Delta \theta^t}{\eta^2}
\label{eq:4b}
\end{equation}
Substituting, \eqref{eq:4a} and \eqref{eq:4b} in \eqref{eq:3b}, we get,
\begin{equation}
\begin{aligned}
    L(\theta^{t+1}) - L(\theta^{t}) = -\eta \beta \left(-\frac{\Delta \theta^{t+1} - \beta \Delta \theta^t}{\eta^2}\right)^T(\Delta \theta)^t  - \eta \left\| -\frac{\Delta \theta^{t+1} - \beta \Delta \theta^t}{\eta^2} \right\|^2
\end{aligned}    
\end{equation}
Ignoring the constants $\beta, \eta$, and replacing them with $c_1, c_2, c_3$,
\begin{equation}
    (\Delta L)^{t+1} = c_1 \|\Delta \theta^{t+1}\|^2 + c_2 \|\Delta \theta^{t}\|^2 + c_3 (\Delta \theta^{t})^T \Delta \theta^{t+1}
\end{equation}
where $c_1, c_2, c_3$ are some fixed functions of $\eta, \beta$. Using the same principle above, and replacing the coefficients with parameter specific $\gamma_k$. However, note that here we have three possibly independent features for each parameter $\theta_k$ - Time difference at $t=t_0$, Time difference at $t=t_1$, and the cross feature between the differences. Hence instead of using a single parameter $\gamma_k$ (as in the case of vanilla gradient descent), one needs to use $(\gamma_{k,0},\gamma_{k,1},\gamma_{k,1})$. The model then becomes,
\begin{equation}
\begin{aligned}
    (\Delta L)^{t+1} = \sum_k \gamma_{k,0} (\Delta\theta_k^{t+1})^2 + & \gamma_{k,1}(\Delta\theta_k^{t})^2 +  \gamma_{k,2}(\Delta\theta_k^{t})(\Delta\theta_k^{t+1})
\end{aligned}    
    \label{eq:momentumsgdmodel}
\end{equation}
Here we consider the parameter to be not-important if $\gamma_{k,0}=\gamma_{k,2}=\gamma_{k,2}=0$, i.e all the coefficients should be irrelevant.

We also replace the causal pruning step, with the following - Starting at $t=2$, compute first-order differences,
    \begin{equation}
        \begin{aligned}
            \Delta \theta^{t,0} &=\theta^{t} - \theta^{t-1} \\
            \Delta \theta^{t,1} &=\theta^{t-1} - \theta^{t-2} \\
            \Delta L^{t} &= L^{t} - L^{t-1}
        \end{aligned}
    \end{equation}
    and fit the model using $\{\Delta \theta^{t,0},\Delta \theta^{t,1},\Delta L^{t}\}_{t=1}^{T}$ as the dataset. Here $\Delta \theta^{0},\Delta \theta^{1}$ denotes the independent variables and $\Delta L$ denotes the dependent variables.
    \begin{equation}
    \begin{aligned}
        \Delta L = \sum_{k} \gamma_{k,0}(\Delta \theta^{0}_{k})^{2} + \gamma_{k,1}(\Delta \theta^{1}_{k})^{2} + \gamma_{k,2}(\Delta \theta^{0}_{k})(\Delta \theta^{1}_{k}) + \alpha \sum_k |\gamma_{k,0}| + |\gamma_{k,1}| + |\gamma_{k,2}|
    \end{aligned}        
    \end{equation}
    We obtain the pruning masks as,
    \begin{equation}
        \{\theta_k \mid \gamma_{k,0} \neq 0 \text{ or } \gamma_{k,1} \neq 0 \text{ or } \gamma_{k,2} \neq 0\}
    \end{equation}

\paragraph{Remark: What about the case with varying learning rates?} It turns out that it is not easy to adapt the above procedure to gradient descent with adaptive learning rates such as ADAM \citep{DBLP:journals/corr/KingmaB14} or RMSProp \citep{rmsprop}. Specifically, one would have to consider all the significant updates till time $t$ - $\{\Delta \theta^t\}_{t=0}^{t=t}$. Since this is computationally expensive and also since it is known that SGD with momentum works sufficiently well in practice \citep{DBLP:conf/iclr/LoshchilovH17}, we do not consider this case in the current scope of the article.

\section{Details of Flat Minima Experiments}
\label{sec:details_flat_mimima}

\begin{figure*}[!t]
\subfigure[Causal Pruning][]{%
    \label{fig:a1a}
    \includegraphics[width=0.3\linewidth]{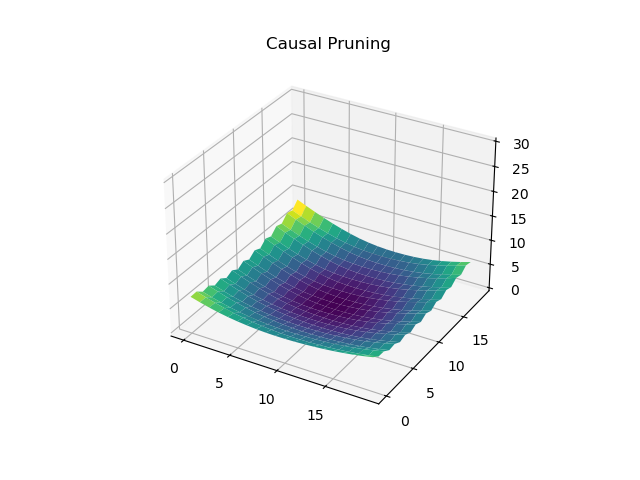}    
}%
\hfill%
\subfigure[Magnitude Pruning][]{%
    \label{fig:a1b}
    \includegraphics[width=0.3\linewidth]{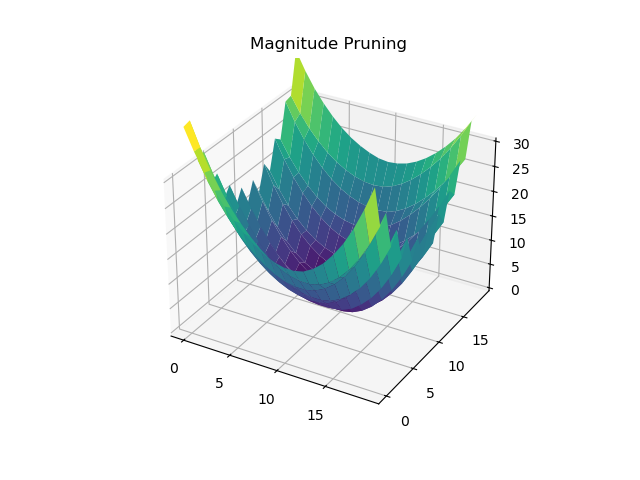}
}%
\hfill %
\subfigure[No Pruning][]{%
    \label{fig:a1c}
    \includegraphics[width=0.3\linewidth]{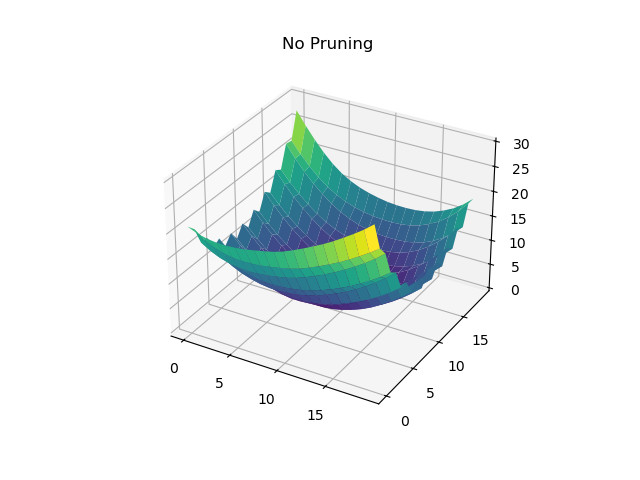}    
}
\caption{Loss Landscape}
\label{fig:a1}
\end{figure*}

To obtain the top eigenvalues from trained networks, we use the stochastic power-iteration method from \citet{DBLP:conf/bigdataconf/YaoGKM20}. For visualizing the minima, we use the method proposed in \citet{DBLP:conf/nips/Li0TSG18}, which is the following:
\begin{itemize}
    \item[1.] We consider two arbitrary directions by initializing a network with weights from a normal distribution with mean 0 and standard deviation 1 - $u_1$ and $u_2$. We then normalize the filters to have the same norm as the original network
    \begin{equation}
        u_{i,j}= \frac{u_{i,j}}{\|u_{i,j}\|} \|d_{i,j}\|
    \end{equation}
    where $u_{i,j}$ refers to the $j^{th}$ filter from $i^{th}$ layer of randomly initialized network, and $d_{i,j}$ refers to the the $j^{th}$ filter from $i^{th}$ layer of the trained network.
    \item[2.] We then plot the loss obtained by $f(\theta^*+\alpha u_1 + \beta u_2)$ where $f$ refers to the network architecture, $\theta^*$ refers to the parameters of the original network and $\alpha, \beta \in (-1,1)$. This is visualized as a 3d plot as shown in the figures. Further, we also scale the z-axis as $\log(1 + f(\theta))$ for better visualization.
\end{itemize}

\section{Experimental Setup}
All experiments were carried out on an NVIDIA DGX V100 server. The server is equipped with an Intel(R) Xeon(R) CPU E5-2698 v4 CPU with 20 physical cores running 40 threads, paired with 256 GB of system memory. The server also houses four NVIDIA Tesla-V100-DGXS GPUs, each possessing 32GB of VRAM. 

We used Adam optimizer for prepruning training and post-pruning training steps with a learning rate of \num{5e-4}. We used the SGD optimizer with a learning rate of \num{1e-3} for the pruning steps. Models were trained until the training loss stopped decreasing beyond the minimum training loss achieved for at least 5 epochs where the current training loss would be considered worse than the best training loss if it didn't improve upon the best loss by more than \num{1e-4}. We took the best model measured by accuracy on the validation set to report the results.

We used $N_{pre} = 10$, $N_{iter} = 10$, $N_{prune} = 10$, and $N_{post} = 200$ unless specified otherwise. We used $batch\_size = 512$ except for \texttt{ResNet50} where we used $batch\_size = 256$ to make the model fit in VRAM. 

\begin{table*}[h]
\caption{Hyperparameters used to run experiments}
\centering
\begin{tabular}{cccc}
{\makecell{Model}} & Dataset & \texttt{L1\_coeff} & \texttt{mag\_prune\_frac} \\
\toprule
\multirow{3}{*}{\makecell{LeNet}} & FashionMNIST & \num{1e-14} & \num{0.1}  \\
& CIFAR10 & \num{1e-14} & \num{0.275} \\
& TinyImageNet & \num{1e-14} & \num{0.275}  \\
\midrule
\multirow{3}{*}{\makecell{AlexNet}} & FashionMNIST & \num{1e-17} & \num{0.275}  \\
& CIFAR10 & \num{1e-16} & \num{0.275} \\
& TinyImageNet & \num{1e-16} & \num{0.196}  \\
\midrule
\multirow{3}{*}{\makecell{ResNet18}} & FashionMNIST & \num{1e-16} & \num{0.174}  \\
& CIFAR10 & \num{1e-15} & \num{0.215} \\
& TinyImageNet & \num{1e-15} & \num{0.14}  \\
\bottomrule
\end{tabular}
\label{table:3}
\end{table*}

Table\ref{table:3} shows the \texttt{L1\_coeff} and \texttt{mag\_prune\_frac} used for experiments reported in this paper. \texttt{L1\_coeff} controls the amount of Causal Pruning in each pruning iteration. \texttt{mag\_prune\_frac} is the fraction of global params pruned in every pruning iteration of Magnitude Pruning. We chose specific \texttt{mag\_prune\_frac} to match the final prune percentage obtained by Causal Pruning.

\end{document}